# FULL OBJECT BOUNDARY DETECTION BY APPLYING SCALE INVARIANT FEATURES IN A REGION MERGING SEGMENTATION ALGORITHM

Reza Oji and Farshad Tajeripour

Department of Computer Engineering and IT, Shiraz University
Shiraz, Iran
`oji.reza@gmail.com`
`tajeri@shirazu.ac.ir`

## ABSTRACT

*Object detection is a fundamental task in computer vision and has many applications in image processing. This paper proposes a new approach for object detection by applying scale invariant feature transform (SIFT) in an automatic segmentation algorithm. SIFT is an invariant algorithm respect to scale, translation and rotation. The features are very distinct and provide stable keypoints that can be used for matching an object in different images. At first, an object is trained with different aspects for finding best keypoints. The object can be recognized in the other images by using achieved keypoints. Then, a robust segmentation algorithm is used to detect the object with full boundary based on SIFT keypoints. In segmentation algorithm, a merging role is defined to merge the regions in image with the assistance of keypoints. The results show that the proposed approach is reliable for object detection and can extract object boundary well.*

## 1. INTRODUCTION

Finding the points with high information namely keypoints from an image is very important and valuable for many applications in computer vision and image processing like object detection, shape recognition and image registation. We can use the keypoints for finding objects in the other images. Object recognition by using the keypoints is very accurate because if the keypoints are identified correctly, they achieve the best information from the image. SIFT [1] is an invariant method with respect to scale, translation, rotation and partially invariant to change in illumination and camera viewpoint. Therefore, the features are very robust for regocnizing an object in different images and detect it by using a poweful segmentation algorithm.

Lowe has presented an object recognition algorithm [2] based on SIFT. Also, many object recognition and detection methods such as [3,4,5] have been presented that each having its own specifications. Some of these algorithm are iterative and based on region merging [6,7]. Interactive image segmentation by region merging based on maximal similarty [8] is a powerful algorithm for detecting object and its boundary, but, it has a defect. In this algorithm the users must indicate some of locations of the background and object to run the algorithm, therefore, this is not an automatic algorithm.

In this paper, we applied SIFT keypoints in a region merging algorithm to recognize the objects in the images and detect them with full boundary. The presented algorithm does not have the stated problem in region merging algorithm. It means that the algorithm does not need to marked regions by user for running. We use the best keypoints of an object which has been obtained from SIFT results and apply them into the test images. Therefore, the method is an automatic algorithm and the achieved keypoints from SIFT have been replaced with user marks. The proposed method, is a fully object boundary detection approach by applying SIFT keypoints in a region merging segmentation algorithm. The keypoints are very efficient and provide best information of objects to identify them in the other images.

## 2. SIFT ALGORITHM OVERVIEW

SIFT has presented by Lowe in 1999 that is an algorithm to detect and describe local features in images. The features are invariant to scale, translation and rotation which can achieve keypoints of image. By using the keypoints, objects in an image can recognize and identify in the other images. major steps of the algorithm is described below.

### 2.1. Scale Space Extrema Detection

The first step is extracting the keypoints that are invariant to changes of scale. Therefore, we need to search for stable features in all possible changes. In previous research it has been showed that for this task, the Gaussian function is the only possible scale-space kernel.

A function, $L(x,y,\sigma)$, is the scale-space of an image which is obtained from the convolution of an input image, $I(x,y)$, with a variable scale-space Gaussian function, $G(x,y,\sigma)$. To efficiently detect stable keypoint locations, Lowe has proposed using the scale-space extrema in difference-of-Gaussian (DoG) function, $D(x,y,\sigma)$. Extrema computed by the difference of two nearby scales separated by a constant factor K (1):

$$\begin{aligned} D(x,y,\sigma) &= (G(x,y,k\sigma) - G(x,y,\sigma)) * I(x,y) \\ &= L(x,y,k\sigma) - L(x,y,\sigma) \end{aligned} \quad (1)$$

This process will be repeated in several octaves. In each octave, the initial image is repeatedly convolved with the Gaussian function to produce the set of scale space images. Adjacent Gaussian images are subtracted to produce the difference-of-Gaussian images. After each octave, the Gaussian images are down-sampled by a factor of 2, and the process is repeated. Figure 1, shows a visual representation from this process.

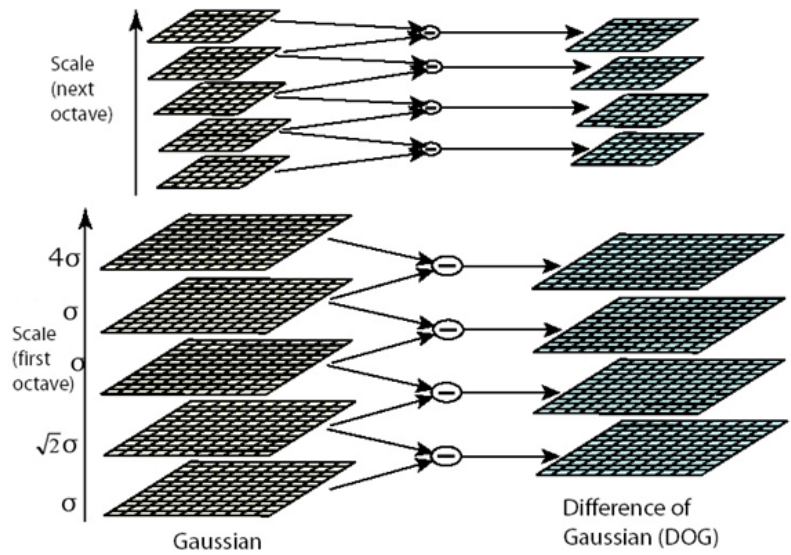

Figure 1. Visual representation of DOG in octaves

Then each sample point (pixel) is compared with its neighbors according to their intensities for finding out whether is smaller or larger than neighbors. For more accuracy, each pixel will be checked with the eight closest neighbors in image location and nine neighbors in the scale above and below (Figure 2). If the point is an extrema against all 26 neighbors, is selected as candidate keypoint. The cost of this comparison is reasonably low because most sample point will be eliminated at first few check.

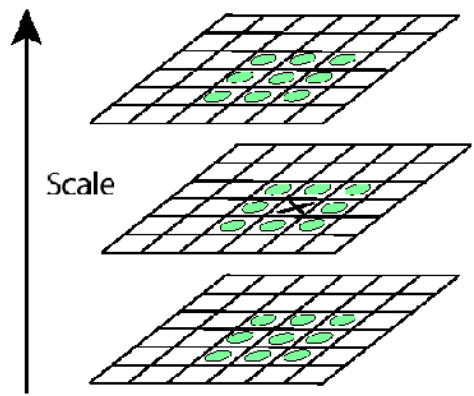

Figure 2. Local extrema detection of DOG

## 2.2. Accurate Keypoint Localization

For each candidate keypoint, interpolation of nearby data is used to accurately determine its position and then too many keypoints that are unstable and sensitive to noise like the points with low contrast and the points on the edge, will be eliminated.

## 2.3. Assigning an Orientation

The next step is assigning an orientation for each keypoint. By this step, the keypoint descriptor [9] can be represented relative to this orientation and therefore obtain invariance to image rotation. For each Gaussian smoothed image sample, the points in regions around keypoint are selected and magnitude, m, and orientations, , of gradiant are calculated (2):

$$m(x, y) = \sqrt{(L(x+1, y) - L(x-1, y))^2 + (L(x, y+1) - L(x, y-1))^2}$$

$$\theta(x, y) = \tan^{-1}((L(x, y+1) - L(x, y-1)) / (L(x+1, y) - L(x-1, y)))$$

(2)

Then created weighted histogram of local gradient directions computed at selected scale. Histogram is formed by quantizing the orientations into 36 bin to covering 360 degree range of orinetations. The highest peak in the histogram is detected where peaks in the orientation histogram correspond to dominant directions of local gradiant. Scale and orientation of the keypoints and eliminating of unstable keypoints are shown in Figure3.

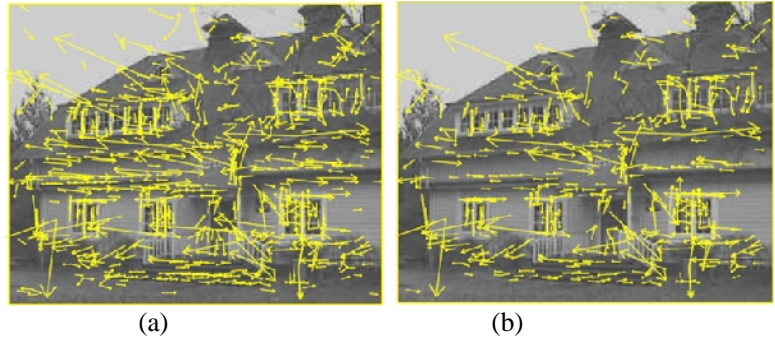

(a)          (b)

Figure 3. (a): scale and orientation of the keypoints, (b): the keypoints after unstable points eliminating

## 2.4. Keypoint Descriptor

The previous operations have assigned location, scale, and orientation to each keypoint and provided invariance to these parameters. Remaining goals are to define a keypoint descriptor for the local image regions and reduce the effects of illumination changes. The descriptor is based on 16×16 samples that the keypoint is in the center of. Samples are divided into 4×4 subregions in 8 direction around the keypoint. Magnitude of each point is weighted and gives less weight to gradients far from keypoint (Figure 4). Therefore, feature vector dimantional is 128 (4×4×8).

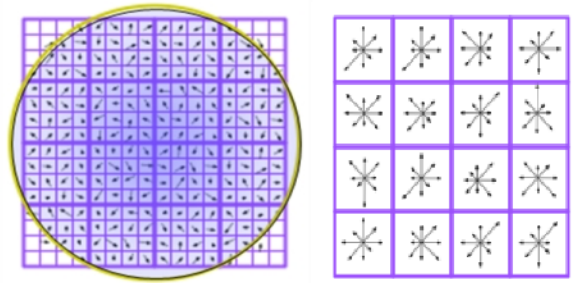

Figure 4. A keypoint descriptor

Finally, vector normalization is applied. The vector is normalized to unit length. A change in image contrast in which each pixel value is multiplied by a constant will multiply gradients by the same constant. Contrast change will be canceled by vector normalization and brightness change in which a constant is added to each image pixel will not affect the gradient values, as they are computed from pixel differences.

Now, we can find the keypoints of an image in the other image and match them together. One image is the training sample of what we are looking for and the other image is the world picture that might contain instances of the training sample. Both images have features associated with them across different octaves. Keypoints in all octaves in one image independently match with all keypoints in all octaves in other image. Features will be matched by using nearest neighbor algorithm. The nearest neighbor is defined as the keypoint with minimum Eculidean distance for the invariant descriptor vector as described upside. Also to solve the problem of features that have no correct match due to some reason like, background noise or clutter, a threshold at 0.8 is chosen for ratio of closest nearest neighbor with second closest nearest neighbor that obtained experimentally . If the distance is more than 0.8, then the algorithm does not match the keypoints together.

## 3. PROPOSED OBJECT DETECTION ALGORITHM

For each object, we train it by SIFT algorithm in several images with different scales and viewpoints to find best keypoints to recognize desire object in the other images. Then, by using these keypoints and region merging algorithm detect the object. A maximal similarity region merging segmentation algorithm was propoesd by Ning in 2010. This algorithm is based on a region merging method with using initial mean shift segmentation and user markers. The user marks some parts of object and background to help the region merging process based on maximal similarity. The non-marker background regions will be automatically merged while the non-marker object regions will be identified and avoided from being merged with background. This algorithm can extract the object from complex scenes but there is a weakness which the process needs user marks, hence, the algorithm is not automatic. In our algorithm, the base of this region merging process is used but with another similarity measure and without user marks. We propose an efficient method for object recognition and detection by applying SIFT keypoints (instead of user marks) to an automatic segmentation based on region merging which can detect the object with full boundary.

### 3.1. Initial Mean Shift Segmentation

An initial segmentation is required to partition image into regions for regions merging in further steps. The mean shift segmentation is used by its software namely the EDISON System [10] for initial segmentation. Mean shift segmentation can preserve the boundries well and has high speed but any other low level segmentation like super-pixel [11] can be used. Please refer to [12,13] for more information about mean shift segmentation. Mean shift segmentation of an image is shown in figure [5].

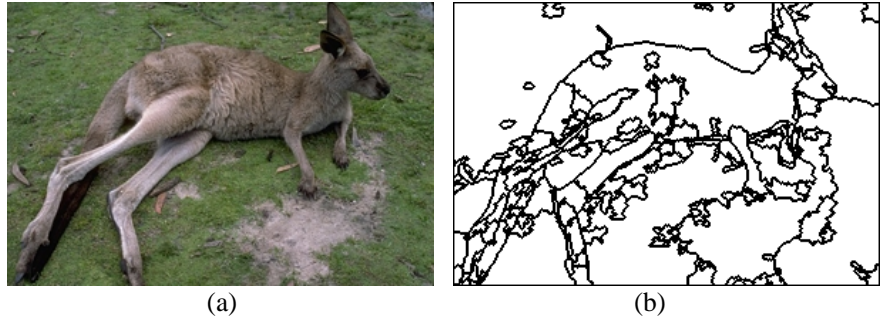

(a) (b)

Figure 5. (a) original image, (b): mean shift segmentation of (a)

## 3.2. Region Merging Based on a Similarity Measure

There is many small region after initial mean shift segmentation. These regions represent using color histogram that is an effective descriptor because different regions from the same object often have high similarity in color whereas there have variation in other aspects like size and shape.

The color histogram computed by RGB color space. Each color chanel is uniformly quantized into 8 levels. Therefore, the feature space is 8×8×8=512 and the histogram of each region is computed in 512 bins. The normalized histogram of a region x denote by $Hist_x$. Now we want to merge the regions by their color histograms that extract the desired object.

When we applied SIFT keypoints in the image, some regions in image that are relevant to object are compoesd of keypoints and the others not. Here, an important issue is how to define the similarity measure between regions which include keypoints and regions without any point so that, the similar regions can be merged. We use a well known goodness of fit statistical metrics, City Block distance, as similarity measure. (X,Y) is a similarity measure between two regions x and y, based on the City Block distance [3].

$$\Delta(X,Y) = \sum_{m=1}^{512} \left| Hist_X^m - Hist_Y^m \right| \qquad (3)$$

Where $Hist_X$ and $Hist_Y$ are the normalized histograms of X and Y. Also, The superscript m is $m_{th}$ member of the histograms. lower City Block distance between x and y means that the similarity between them is higher.

## 3.3. Region Merging Process

In the merging process, we have three regions in the image with diferent lables. The object regions denote by $R_O$ that are identified by obtained keypoints from training SIFT algorithm. The regions around the images denote by $R_B$ that usually are not inclusive objects, therefore, we cover all around the image as initial background regions to help merging process start. The third regions are regions without any sign and denote by N.

A merging rule is defined in continue (4). Let Y be an adjacent region of X and $S_Y$ is the set of Y's adjacent region that X is one of them. The similarity between Y and $S_Y$, i.e. (Y,$S_Y$), is calculated. X and Y will be merged together if and only if the similarity between them is the maximal among all the similarities (Y,$S_Y$).

International Journal of Artificial Intelligence & Applications (IJAIA), Vol.3, No.5, September 2012

Merge X and Y if    (X,Y) = max    (Y,S_Y)                                                                             (4)

The main strategy is to keep object regions from merging and merge background regions as many as possible. In the first stage we try to merge background regions with their adjacent regions. Each region B ϵ $R_B$ will be merged into adjacent region if the merging rule is be satisfied. If the merging accured, the new region has the same label as region B. This stage will be repeated iteratively and in each iteration $R_B$ and N will be updated. Obviously, $R_B$ expands and N shrink due to merging process. The process in this stage stops when background regions $R_B$ can not find any region for new merging.

After this stage, some background regions are still, that can not be merged due to merging rule. In the next stage we will focus on the remaining regions in N from the first stage which are combination of background (N) and object ($R_O$). As before, the regions will be merged based on merging rule and the stage will be repeated iteratively and updated and stops when the entire region N can not find new region for merging. These stages will be repeated again until the merging is not occured. Finaly, all remaining regions in N will be labeled as object and merged into $R_O$ and we can easily extract the object from the image.

## 4. EXPERIMENTAL RESULTS

Our results show that the proposed method is powerfull and robust in object recognition and detection. We used an online dataset [14] and trained three objects of it ( book, monitor and stapler)  in several views with different scales and illuminations for checking our method. Using SIFT keypoints in region merging algorithm based on maximal similarity, helped us to obtain significant results, because these are very exact and come from a strong algorithm (SIFT) with statistical basis. Moreover, the keypoints give the best information of objects which are very useful for merging process. Figure 6, shows SIFT keypoints on the images with initial mean shift segmentation  and detected objects with their boundaries. First row, (a), is relevant to the book dataset, second  row, (b), is the monitor and thirth row, (c), is the stapler.

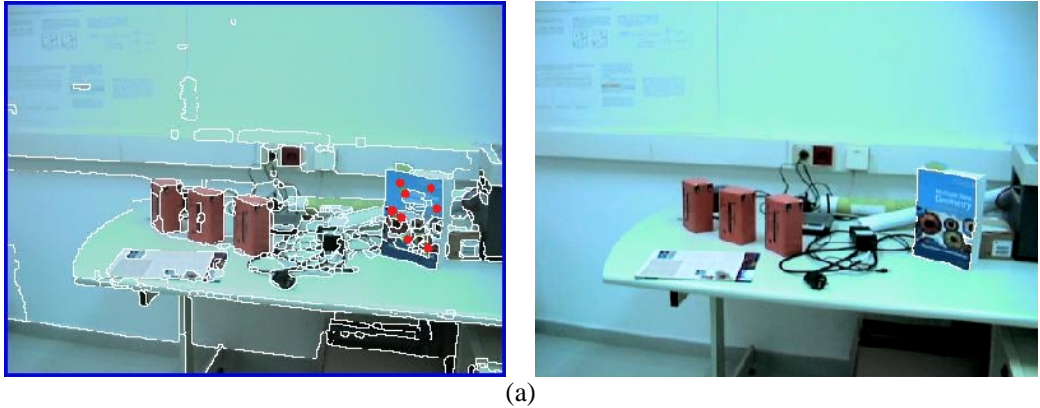

(a)

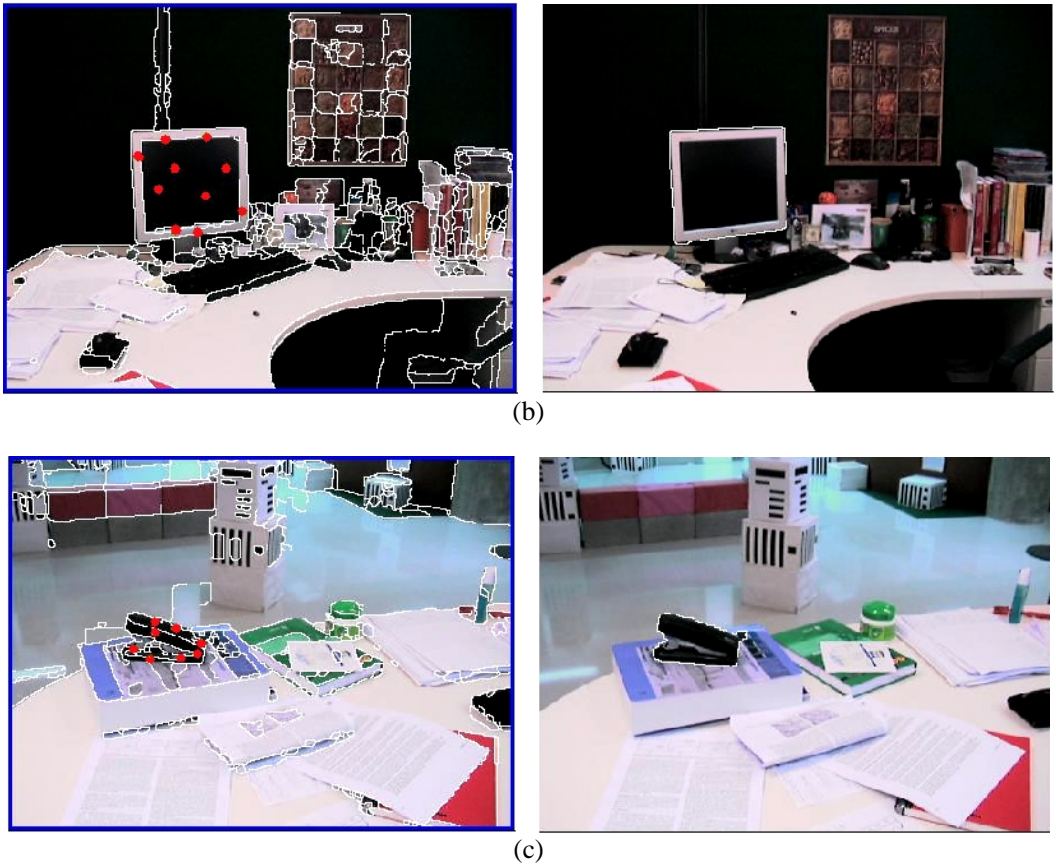

Figure 6. Left column: SIFT keypoints with initial mean shift segmentation. right column: detected object with its boundary.

Accuracy rate (5) of full boundary detection for each object has been computed and shown in Table 1. Results show that the accuracy rate of datasets are close together.

$$Accuracy\ Rate = 100 \times \frac{N_{to}}{N_{do}} \qquad (5)$$

Where $N_{to}$ indicates the number of trained images of each dataset and $N_{do}$ means the number of full detected objects respective to the dataset.

Table 1. Accuracy rate of three datasets.

| Dataset | No. of trained images | No. of test images | No. of full detected object | Accuracy rate |
| --- | --- | --- | --- | --- |
| Book | 48 | 24 | 22 | 91.66 % |
| Monitor | 48 | 24 | 21 | 87.5 % |
| Stapler | 48 | 24 | 20 | 83.33 % |

In continuation, some of the other experimental results are shown in Figure 7.

International Journal of Artificial Intelligence & Applications (IJAIA), Vol.3, No.5, September 2012

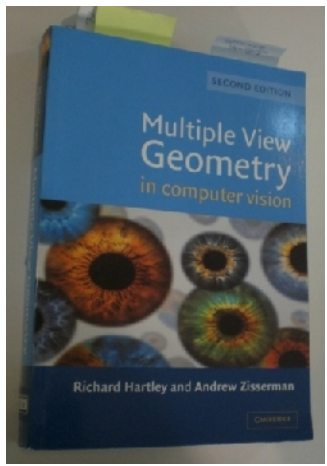
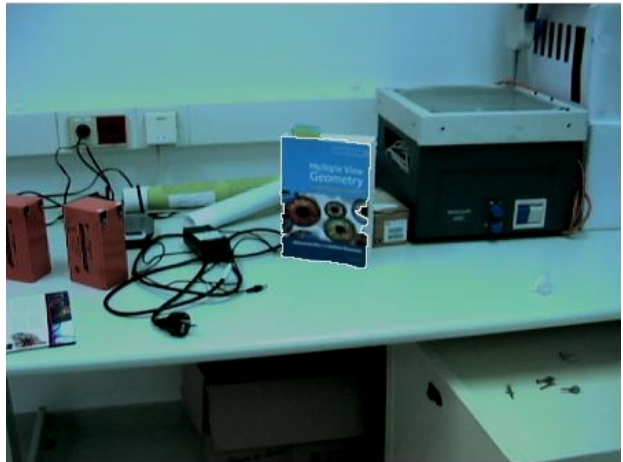

(a)

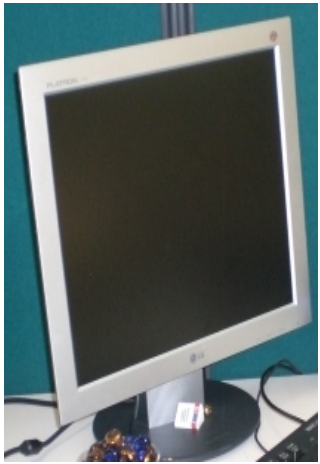
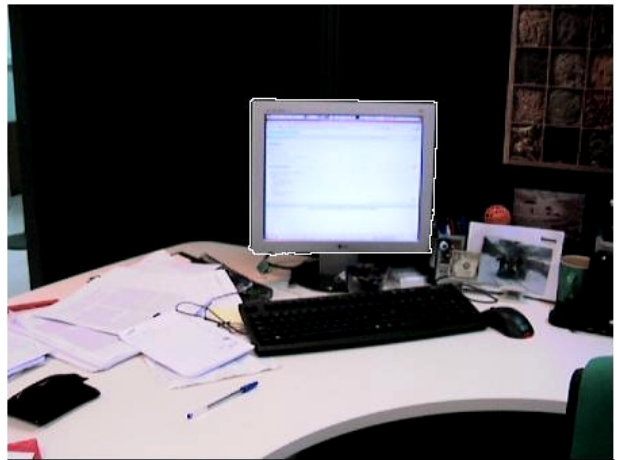

(b)

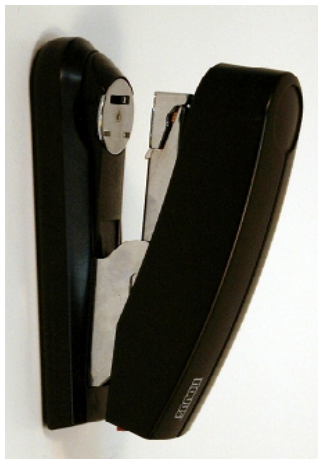
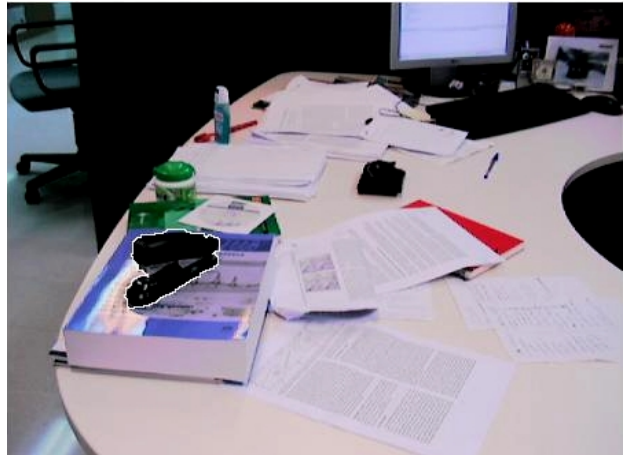

(c)

Figure 7. Left column: one of the trained images for an object, right column: the detected object. (a): book, (b): monitor (c): stapler.

## 5. CONCLUSIONS AND FUTURE WORK

This paper proposed a method for full object boundary detection. This goal is obtained by combining SIFT and a region merging algorithm based on similarity measure. We trained different objects seperately by several images with various aspects, scale, illumination and camera viewpoints to find the best keypoints for recognizing them in the other images. Then, these keypoints will be applied to the region merging algorithm that is initially segmented by mean shift segmentation. Region merging algorithm is started by using keypoints and similarity measure (City Block distance). Finaly, the object will be detected well with its boundary. A final conclusion is that the more keypoints are obtained, and the more accurate they are, the results will be better and more acceptable. Currently, we are working to extract the keypoints with a fully affine invariant transform (ASIFT) and cheking other similarity measures in the merging process. We hope to decrease the defect obtained by changes in camera viewpoints compare to SIFT and improve the algorithm performance.